\documentclass[sigchi]{acmart}
\AtBeginDocument{%
  \providecommand\BibTeX{{%
    \normalfont B\kern-0.5em{\scshape i\kern-0.25em b}\kern-0.8em\TeX}}}

\setcopyright{none}
\renewcommand\footnotetextcopyrightpermission[1]{} 
\usepackage{multirow}
\usepackage{graphicx}
\usepackage{subcaption}

\begin{document}

\title{Transformer Networks for Data Augmentation of Human Physical Activity Recognition}

\author{Sandeep Ramachandra}
\affiliation{%
 \institution{Ubiquitous Computing, University of Siegen}
 \streetaddress{H\"olderlinstr. 3}
 \postcode{57076}
}
\email{sandeep.ramachandra@student.uni-siegen.de}
\author{Alexander Hoelzemann}
\affiliation{%
 \institution{Ubiquitous Computing, University of Siegen}
 \streetaddress{H\"olderlinstr. 3}
 \postcode{57076}
}
\email{alexander.hoelzemann@uni-siegen.de}
\author{Kristof Van Laerhoven}
\orcid{0000-0001-5296-5347}
\affiliation{%
 \institution{Ubiquitous Computing, University of Siegen}
 \streetaddress{H\"olderlinstr. 3}
 \postcode{57076}
}
\email{kvl@eti.uni-siegen.de}

\begin{abstract}
Data augmentation is a widely used technique in classification to increase data used in training. It improves generalization and reduces amount of annotated human activity data needed for training which reduces labour and time needed with the dataset. Sensor time-series data, unlike images, cannot be augmented by computationally simple transformation algorithms. State of the art models like Recurrent Generative Adversarial Networks (RGAN) are used to generate realistic synthetic data.  In this paper, transformer based generative adversarial networks which have global attention on data, are compared on PAMAP2 and Real World Human Activity Recognition data sets with RGAN. The newer approach provides improvements in time and savings in computational resources needed for data augmentation than previous approach.
\end{abstract}


\renewcommand{\shortauthors}{Ramachandra, Hoelzemann and Van Laerhoven}
\maketitle
\pagestyle{plain}

\section{Introduction}
Deep learning networks have progressed substantially, both due to improvements in algorithms as well as availability of powerful computing hardware.  These networks, however, require large amounts of data to train well, and such large benchmark datasets are not available for some domain, and human activity recognition from wearable sensors is such a field.
\par
Data augmentation in the context of deep learning, is a technique used to increase the number of input data by adding modified data or by creating artificial data which are similar to real data \cite{eyobu_LSTM_HAR_2018}. Data augmentation acts as a regularizer, leading to generalized neural networks which give better results on unseen data. Image data can for instance be augmented by techniques such as geometric transformations or color space transformations which does not take much computational resources. Timeseries data like the raw output of inertial sensors is harder to augment, since any changes made to timeseries data due to augmentation cannot be visually identified. This is more complicated in human activity datasets since many activities like standing and sitting are very similar and any change may alter what activity the classifier identifies the data with. Annotation of collected data is a labour-intensive task and takes a long time for a sizable dataset \cite{abedin_ClusteringHAR_2020}. The development of Generative Adversarial Networks (GAN) lead to neural networks outputting real-like data from random inputs. This is used to augment limited data. However, GANs do not always converge due to mode collapse arising from alternating gradient descent.\par

This work introduces a GAN architecture that is based on Transformer Networks. We evaluate our architecture with a recently published one \cite{hoelzemann_RGAN_2021} that is based on Long Short-Term Memory (LSTM). LSTMs has the disadvantage that they process incoming data sequentially. However, Transformers process data in a parallel stream which helps to speed up training drastically. These Transformer's parameters grow quadratically with input size and can outstrip available GPU memory. However, for human activity that have small input sizes, this is not a big concern. \par

We validated our network with leave-one-subject-out (LOSO) cross-Validation and on two publicly available datasets, PAMAP2 \cite{reiss2012introducing} and Real World Human Activity Recognition (RWHAR) \cite{RWHAR_dataset}. For the PAMAP2 dataset, we used 7 common activities with 27 sensor channels (3 dimensions each for accelerometer, gyroscope and magnetometer in 3 Inertial Measurement Units). For the RWHAR dataset, the dataset consists of 8 activities with 6 sensor channels (3 dimensions for accelerometer and gyroscope of the smartphone attached to the chest).\par

\section{Related Work}
In sequence to sequence transformations like translating between two languages, transformers outperforms LSTM networks in both resources and in performance metrics \cite{gehring_convolutional_2017}, \cite{vaswani2017attention}. In Human activity recognition, Mahmud et al. \cite{mahmud_human_2020} and Murahari et al. \cite{murahari_HARattention_2018} uses transformers to capture the spatial temporal context from the feature space of sensor reading sequence and classify the sensor data.The transformer networks performs very well in this field as well. This suggests that the use of transformers in data augmentation can be beneficial.  \par

Data augmentation is a topic under research in human activity recognition. Alawneh et al. \cite{alawneh2021enhancing} makes a case that using data augmentation in human activity recognition improves accuracy of trained classifiers. Li et al. \cite{Li_AcitivityGAN_2020} uses a convolutional GAN since Recurrent GANs do not converge consistently, to generate synthetic human activity physical data which were distinguishable by visualization techniques. Hoelzemann et al. \cite{hoelzemann_RGAN_2021} used a state of the art Recurrent GAN (RGAN), a GAN with an LSTM layer, for data augmentation with the resulting artificial data verified to be similar to original data. \par

Transformers in data augmentation has been explored in Zhang et al. \cite{zhang_self-attention_2019} which utilises self attention from transformers to explore long range dependencies in internal representation of images. Another network focusing on synthesizing images is the GANBERT or GAN with a bidirectional encoder representation from transformer \cite{shin_ganbert_2020} which uses self attention to generate difficult medical images like that from MRI and PET scans. \par

\section{Methodology}\label{chapter:experiment}
\begin{figure}[ht]
    \centering
    \includegraphics[width=\columnwidth]{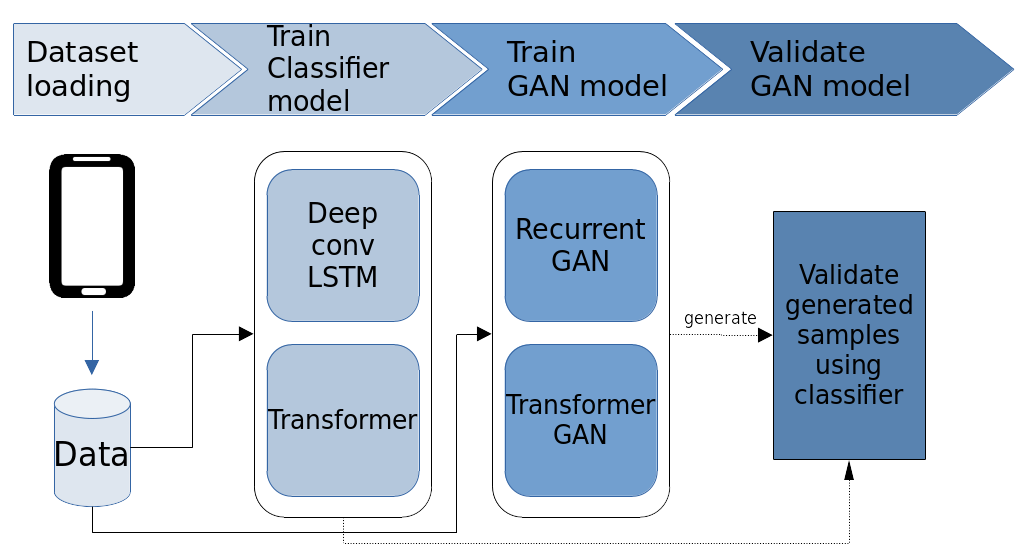}
    \caption{Flowchart of the data augmentation process for human activity recognition.}
    \label{fig:methodology}
\end{figure}
\begin{enumerate}
    \item Ready the dataset for loading : This means that the dataset has been cleaned and loaded up in a format that is ready for the training and validation steps.
    \item Train the validation model : The main objective is to train a model of very high quality to classify signals to classes. The validation model chosen here is one DeepConvLSTM model \cite{Ordonez_deepconvlstm_2016} and one transformer model \cite{wu_deep_2020} for each of the datasets.
    \item Train the GAN models : In this work, an LSTM based Recurrent GAN and a transformer based GAN is compared. New models are trained for each class, so the dataset needs to be separated by the class. 
    \item Validate artificial signal using validation model : The synthesized data from the generator needs to be validated using F1 score from the trained classifier.
\end{enumerate}

\subsection{GAN Models}
\begin{figure}[htbp]
    \centering
    \begin{subfigure}[b]{\columnwidth}
            \centering 
            \includegraphics[width=\textwidth]{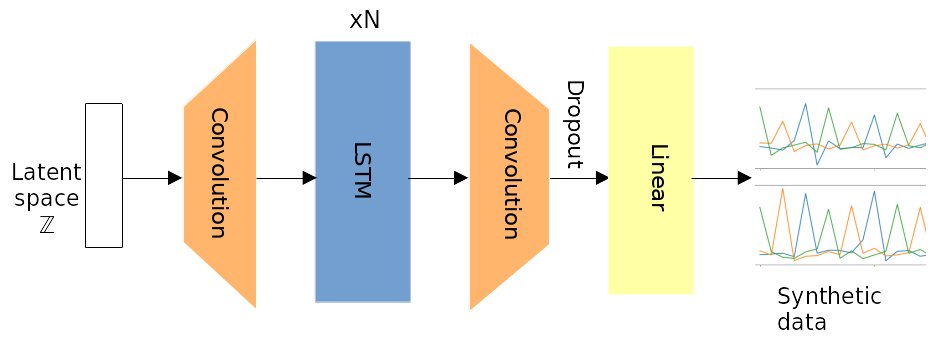}
            \caption[]%
            {{\small RGAN Generator}}    
            \label{fig:RGAN_GEN}
    \end{subfigure}
    \vskip\baselineskip
    \begin{subfigure}[b]{\columnwidth}
            \centering 
            \includegraphics[width=\textwidth]{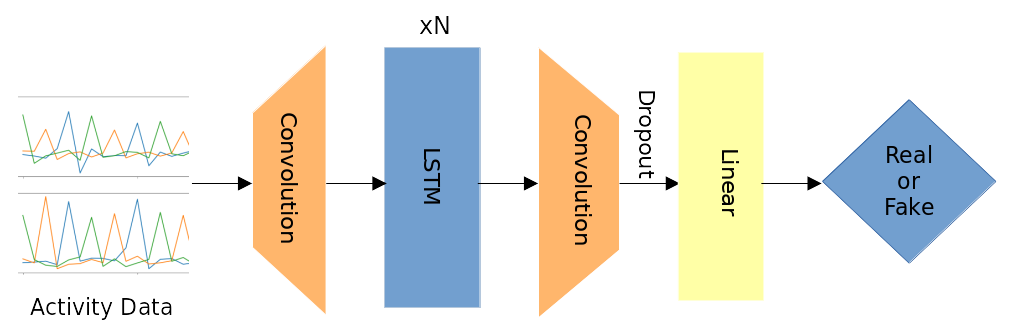}
            \caption[]%
            {{\small RGAN Discriminator}}    
            \label{fig:RGAN_DIS}
    \end{subfigure}
    \caption{Architecture of state of the art Recurrent Generative Adversarial Networks(RGAN). The trapezoid of convolution depict the increase or decrease in channel size based on the divergent or convergent shape of trapezoid respectively.}
    \label{fig:RGAN}
    \vspace{-0.4cm}
\end{figure}

The LSTM based Recurrent GAN (RGAN) (see figure \ref{fig:RGAN}) in both generators and in discriminators, has convolutions going into and coming out of the stacked LSTM layer \cite{hoelzemann_RGAN_2021}. The input to the generator is a 1D vector of random numbers sampled from a standard normal distribution. The input to the discriminator is a vector of shape cxL with c channels and L length (the PAMAP2 dataset has a shape of 27x100 and RWHAR has 6x50). The channels of the input are is progressively increased using 1D convolutions (with padding) to increase the number of trainable parameters in the LSTM as well as increase the receptive field on original data. The stacked LSTM observes the trends in the data sequentially and its hidden state is used to classify the input data to N classes via a fully connected layer. There is a dropout before the fully connected layer providing regularisation during training. The fully connected layer is realised by a 1D convolutional layer. The hyperparameters in this model are the noise length, generator LSTM layers, and discriminator's LSTM's hidden size and layers. \par
\begin{figure}[htbp]
    \centering
    \begin{subfigure}[b]{\columnwidth}
            \centering 
            \includegraphics[width=\textwidth]{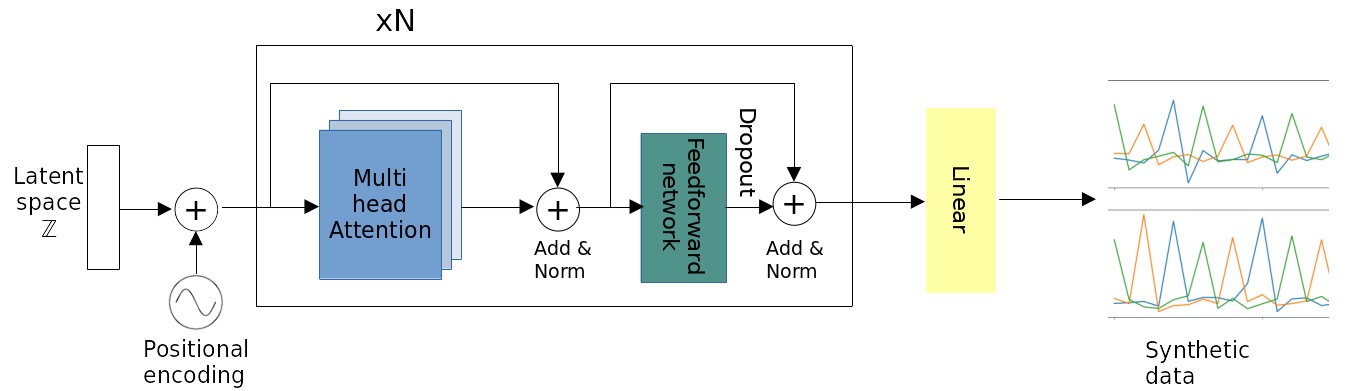}
            \caption[]%
            {{\small TGAN Generator}}    
            \label{fig:TGAN_GEN}
    \end{subfigure}
    \vskip\baselineskip
    \begin{subfigure}[b]{\columnwidth}
            \centering 
            \includegraphics[width=\textwidth]{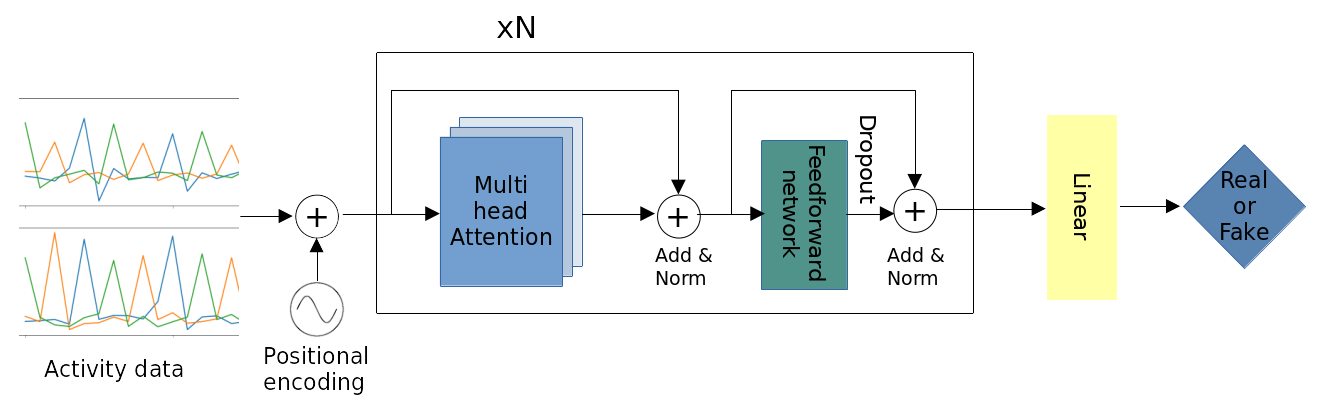}
            \caption[]%
            {{\small TGAN Discriminator}}    
            \label{fig:TGAN_DIS}
    \end{subfigure}
    \caption{Architecture of proposed Transformer Generative Adversarial Networks(TGAN). The attention is realised by a stacked scaled dot product attention \cite{vaswani2017attention}. The feedforward network takes the form of 2 linear layers going from input size to a fixed size and then returning to original input size.}
    \label{fig:TGAN}
    \vspace{-0.3cm}
\end{figure}
The Transformer GAN (TGAN) (see figure \ref{fig:TGAN}) follows a similar approach to the encoder of the traditional transformer encoder layer \cite{vaswani2017attention}. The inputs to both discriminator and generator are same as the ones for Recurrent GAN. The input is not processed sequentially as in Recurrent layers but in parallel. To give the network positional awareness, a regular signal, typically sine/cosine signal, is added to the input. This is then passed to the multi head self attention blocks which is realised by a scaled dot product attention.
\begin{equation}
    Attention (Q, K, V) = softmax(\frac{QK^T}{\sqrt{d_k}})V
    \label{eq:attention}
\end{equation}
The inputs Query(Q), Key(K), and Value(V) are obtained by a 1 to 1 1D convolution of input for each of the three. $d_k$ is a scaling factor for the attention. The multi head attention block are followed by skip connections and then a layer normalization. Then a feedforward network in the shape of two fully connected layers going from input shape to a fixed dimension and then back to input shape. This is followed by another skip connection and layer normalization. This forms the encoder part of the network which is repeated sequentially so the network can learn the features of the input better. Finally, a fully connected layer realised by 1D convolutions with dropout is added so that the intermediary encoding done by the encoder is converted to the synthetic data in case of generator and to real/synthetic classification in case of discriminator. The hyperparameters available are the number of heads of the self attention layer of generator and discriminator, the number of encoder layers stacked in the generator and discriminator, the dimension of feedforward network, and the length of the noise vector. \par 

\section{Results and Evaluation}

\begin{figure}[htb]
    \centering
    \begin{subfigure}[b]{0.49\columnwidth}
            \centering 
            \includegraphics[width=\textwidth]{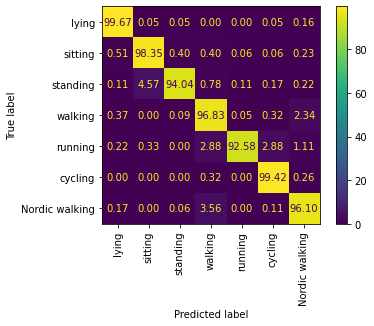}
            \caption[]%
            {{\small PAMAP2 DeepConvLSTM}}    
            \label{fig:ConmfmatPAMAP2LSTM}
    \end{subfigure}
    \begin{subfigure}[b]{0.49\columnwidth}
            \centering 
            \includegraphics[width=\textwidth]{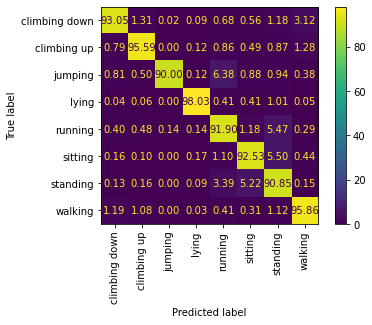}
            \caption[]%
            {{\small RWHAR DeepConvLSTM}}    
            \label{fig:ConfmatRWHARLSTM}
    \end{subfigure}
    \vskip\baselineskip
    \begin{subfigure}[b]{0.49\columnwidth}
            \centering 
            \includegraphics[width=\textwidth]{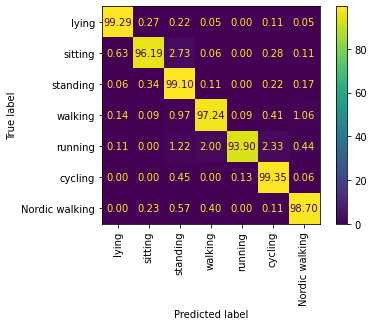}
            \caption[]%
            {{\small PAMAP2 Transformer}}    
            \label{fig:ConmfmatPAMAP2transformer}
    \end{subfigure}
    \begin{subfigure}[b]{0.49\columnwidth}
            \centering 
            \includegraphics[width=\textwidth]{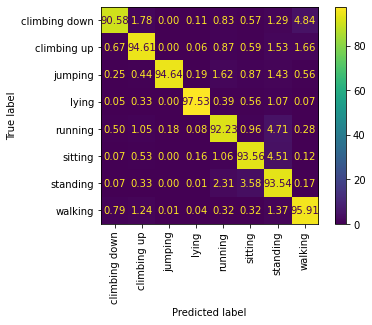}
            \caption[]%
            {{\small RWHAR Transformer}}    
            \label{fig:COnfmatRWHARtransformer}
    \end{subfigure}
    \caption{The confusion matrices in percentages for all trained models and both datasets. The overall validation F1 score of the models are (a) 96.21\% (b) 91.8\% (c) 97\% (d) 92.6\%. Table \ref{tab:GAN_results} shows the sizes of the trained validation models. The transformer classifiers are smaller, faster and have better validation F1 scores than DeepConvLSTM models.}
    \label{fig:ValidationConfmat}
    \vspace{-0.4cm}
\end{figure}
The models are trained with the given hyperparameters chosen by manual search through the configuration space \cite{sandeep_code_2021}. Figure \ref{fig:ValidationConfmat} shows the confusion matrices for the trained models. The Transformer outperforms the DeepConvLSTM in classification problems, both in time and in validation F1 scores (see figure \ref{fig:ValidationConfmat}. In each dataset, the transformer, despite being a smaller model consistently does better than the DeepConvLSTM model. In the classes where it is worse, it still performs very good. The margin of false classifications in those cases is only marginally lower. \par

The Transformer seems like a clear choice from figure \ref{fig:ValidationConfmat} for use in the GAN as a validation model. However in actual use, this is not the case. RGAN appears to prefer DeepConvLSTM models and the Transformer GAN prefers a Transformer classifier models. They give higher validation F1 scores when paired with its preferred model and may not even train to completion with the other model. This may be due to the GAN learning the distribution which can pass the validation. The GANs seem to be able to learn the distribution needed when paired with a similar recurrent based layers i.e the LSTMs of the GAN can learn the distribution needed for DeepConvLSTM but not of the self attention layer and likewise for Transformer GAN. This is a curious observation of the behaviour of GAN, although it is not conclusively proven in this work. This is why there are 4 models trained for 4 data augmentation experiments. \par
\begin{table}[htb]
    \centering
    \begin{tabular}{|c|c|c|c|c|}
        \hline
         Dataset& Model&\begin{tabular}{@{}c@{}}Parameter \\ Size\end{tabular}&\begin{tabular}{@{}c@{}}Augmented \\ Data\end{tabular}&\begin{tabular}{@{}c@{}}Speed up \\ (Time per \\ epoch)\end{tabular}\\
         \hline
         \multirow{6}{*}{PAMAP2}&\multirow{3}{*}{RGAN}&G : 17.6M&\multirow{3}{*}{Yes}&\multirow{3}{*}{\begin{tabular}{@{}c@{}}1x \\ (13 secs)\end{tabular}}\\
         &&D : 697K & &\\
         &&V : 3.00M&& \\
         \cline{2-5}

         &\multirow{3}{*}{TGAN}&G : 765K&\multirow{3}{*}{Yes}&\multirow{3}{*}{\begin{tabular}{@{}c@{}}2.6x \\ (5 secs)\end{tabular}} \\
         &&D : 7.80M & & \\
         &&V : 984K&& \\
         \hline
         \multirow{6}{*}{RWHAR}&\multirow{3}{*}{RGAN}&G : 2.10M&\multirow{3}{*}{No*}&\multirow{3}{*}{\begin{tabular}{@{}c@{}}1x \\ (24 secs)\end{tabular}} \\
         &&D : 1.00M & & \\
         &&V : 11.5M&& \\
         \cline{2-5}
         &\multirow{3}{*}{TGAN}&G : 4.40M&\multirow{3}{*}{Yes}&\multirow{3}{*}{\begin{tabular}{@{}c@{}}3x \\ (8 secs)\end{tabular}} \\
         &&D : 2.20M & & \\
         &&V : 10.3M&& \\
         \hline
    \end{tabular}
    \caption{Table of the performance of trained GAN models.RGAN stands for Recurrent GAN and TGAN stands for Transformer GAN. G is Generator, D is Discriminator, and V is Validation model. For speedup, the RGAN times are used as baseline for calculations for each dataset. The times and speeds shown are the average for the first activity only. The parameter sizes are truncated to first decimal place and are in thousands(K) and millions(M). * - The generated data for all activities in dataset did not satisfy the >95\% validation model F1 score. The validation model training speed for transformer model was 4 to 4.5 times the speed of DeepConvLSTM model.}
    \label{tab:GAN_results}
    \vspace{-0.6cm}
\end{table}
\begin{figure}[htb]
    \centering
    \begin{subfigure}[b]{0.49\columnwidth}
            \centering 
            \includegraphics[width=\textwidth]{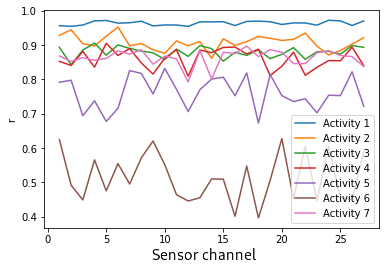}
            \caption[]%
            {{\small  PAMAP2 Recurrent GAN}}    
            \label{fig:CorrPAMAP2LSTM}
    \end{subfigure}
    \hfill
    \begin{subfigure}[b]{0.49\columnwidth}
            \centering 
            \includegraphics[width=\textwidth]{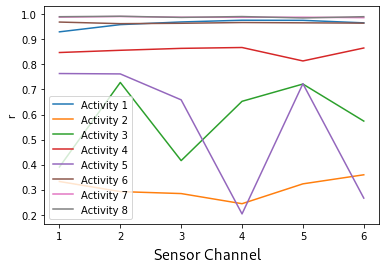}
            \caption[]%
            {{\small  RWHAR Recurrent GAN}}    
            \label{fig:CorrRWHARLSTM}
    \end{subfigure}
    \vskip\baselineskip
    \begin{subfigure}[b]{0.49\columnwidth}
            \centering 
            \includegraphics[width=\textwidth]{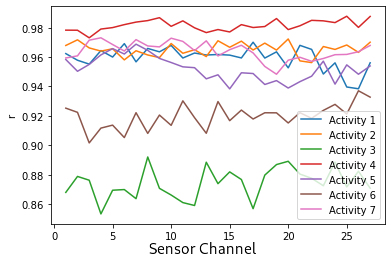}
            \caption[]%
            {{\small PAMAP2 Transformer GAN}}    
            \label{fig:CorrPAMAP2transformer}
    \end{subfigure}
    \hfill
    \begin{subfigure}[b]{0.49\columnwidth}
            \centering 
            \includegraphics[width=\textwidth]{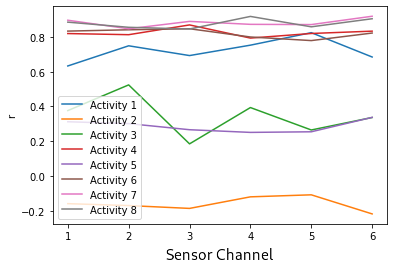}
            \caption[]%
            {{\small  RWHAR Transformer GAN}}    
            \label{fig:CorrRWHARtransformer}
    \end{subfigure}
    \caption{Relation between real data and each model's synthetic data is shown using Pearson correlation coefficient. Mean of 10 samples for each are taken. Each sensor channel of the mean real data is compared to the corresponding sensor channel of the generated data. PAMAP2 has 27 channels - 3 IMUs with 3D Accelerometer, 3D gyroscope and 3D magnetometer. RWHAR has 6 channels - 1 IMU with 3D accelrometer and 3D gyroscope.}
    \label{fig:Corrcoeff}
    \vspace{-0.4cm}
\end{figure}
Table \ref{tab:GAN_results} shows the performances of the 4 GANs which have been trained. The table lists the parameter size for the first try of the GAN training for generator and discriminator. The GAN models are so large as they were not optimized with a hyperparameter search as they have very large training time. The given hyperparameters are the first ones which made the GAN network to converge. The times per epoch is the average time taken for the model to complete the training step and validation step. To calculate the speed up on using the model, the time of the baseline model is divided by time taken by the model. The advantage of using the transformer is immediately apparent. The time taken for each epoch for the Transformer GAN in spite of being bigger than Recurrent GAN, is much smaller than Recurrent GAN. All but one of the models trained the augmented data for all activities successfully. For the Recurrent GAN trained on RWHAR dataset, the GAN produced augmented data which produced 91\% F1 score on the validation model which is still a respectable output. \par

The synthetic data generated from each GAN looks different from the real data but since the shown samples are just one of many real samples, it is difficult to make any conclusions regarding the data. So the mean of 'X' samples of real data and the mean of 'X' samples of syntheticdata are used in the calculation of Pearson correlation coefficient to make a simple comparison. The coefficient is a measure of the relationship between the two variables. The formula is 
\begin{equation}
    r = \frac{N\sum xy - (\sum x)(\sum y)}{\sqrt{[N\sum x^2 - (\sum x)^2][N\sum y^2 - (\sum y)^2]}}
\end{equation}
where r is the Pearson correlation coefficient of the two variables x and y, and N is the number of pairs of scores in x and y. The coefficient is scaled from -1 to 1 where 1 denotes that the two variables are proportional to each other and -1 denotes inverse proportionality. 0 means the two variables are not related to one another. Ideally, very closely related data like a real sample and perfect syntheticdata should have a coefficient of 1. \par
 
 Each axes of the data is compared to one another using the formula and the resulting coefficients are plotted for each activity. The figure \ref{fig:Corrcoeff} shows the graphs for the 4 dataset-model combinations. Any correlation of 0.8 and above correlates very well for use in data augmentation. This means that for PAMAP2, the Transformer GAN works splendidly with all activities of good quality, far better than Recurrent GAN in all activities. However there appears to be some issues with the RWHAR dataset for both models. Recurrent GAN fails in 3 activities and Transformer GAN in 4 activities. Activity 2(climbing down) is  appears to be problematic since both have below 0.2 and transformer appears to be negatively correlated. This seems to be linked to one of the subjects of the dataset who wore their sensor upside down which has had a detrimental effect on both models but the transformer GAN is more severely afflicted. This suggests that the Transformer GAN is more sensitive to outlying data. \par

\section{Conclusions and outlook}
 This paper proposed a Transformer-based GAN for data augmentation of human activity data, in order to provide a speedup benefit over the existing Recurrent GAN and a performance benefit over a Convolutional GAN. This GAN is used along with a validation model which is used to verify that the output of the GAN is identifiable. The Transformer GAN is compared with Recurrent GAN using two public datasets, PAMAP2 and Real World Human Activity Recognition. In validation models, Transformer model outperformed DeepConvLSTM model in both F1 scores and in model parameter sizes. The Transformer GAN provides a 2 - 3 times speedup over Recurrent GAN in both datasets.  The performance of the Transformer GAN is as good as if not better than the Recurrent GAN though it appears to be more sensitive to errant input data than Recurrent GAN. For that reason we are able to state that we (1) implemented the model and it generates data as needed for training (2) beat current benchmarks for data augmentation of human activity data with respect to the GAN training time (3) pointed out the importance to ensure that the produced data is variable but still belong to the original distribution of the input data and therefore recognizable by the classification network. \par

\begin{acks}
We thank our anonymous reviewers for their many insightful suggestions and feedback.
\end{acks}

\bibliographystyle{ACM-Reference-Format}
\bibliography{source}

\end{document}